%% file: sn-article.tex
\documentclass[sn-mathphys-num]{sn-jnl}

\usepackage{graphicx}%
\usepackage{multirow}%
\usepackage{amsmath,amssymb,amsfonts}%
\usepackage{amsthm}%
\usepackage{mathrsfs}%
\usepackage[title]{appendix}%
\usepackage{xcolor}%
\usepackage{textcomp}%
\usepackage{manyfoot}%
\usepackage{booktabs}%
\usepackage{algorithm}%
\usepackage{algorithmicx}%
\usepackage{algpseudocode}%
\usepackage{listings}%

\usepackage{algorithm}
\usepackage{algpseudocode}
\usepackage{amsmath}   
\usepackage{xcolor} 
\usepackage{breakurl}
\RequirePackage[none]{hyphenat}

\usepackage{makecell}
\usepackage{url}
\usepackage{adjustbox}
\usepackage{caption}
\usepackage{subcaption}
\usepackage{soul}
\usepackage{natbib}
\usepackage{nomencl}
\usepackage{xurl}
\makenomenclature

\theoremstyle{thmstyleone}%
\theoremstyle{thmstyletwo}%

\theoremstyle{thmstylethree}%
\hypersetup{breaklinks=true}

\usepackage{enumitem} 

\begin{document}
\sloppy

\title[Article]{XAI-based Feature Ensemble for Enhanced
Anomaly Detection in Autonomous Driving
Systems}


\author[1]{\fnm{Sazid} \sur{Nazat}}\email{snazat@purdue.com}

\author*[2]{\fnm{Mustafa} \sur{Abdallah}}\email{abdalla0@purdue.edu}


\affil[1]{\orgdiv{Electrical and Computer Engineering Department}, \orgname{Purdue University in Indianapolis}, \orgaddress{\street{420 University Blvd}, \city{Indianapolis}, \postcode{46202}, \state{Indiana}, \country{USA}}}

\affil*[2]{\orgdiv{Computer and Information Technology Department}, \orgname{Purdue University in Indianapolis}, \orgaddress{\street{420 University Blvd}, \city{Indianapolis}, \postcode{46202}, \state{Indiana}, \country{USA}}}



\abstract{The rapid advancement of autonomous vehicle (AV) technology has introduced significant challenges in ensuring transportation security and reliability. Traditional AI models for anomaly detection in AVs are often opaque, posing difficulties in understanding and trusting their decision-making processes. This paper proposes a novel feature ensemble framework that integrates multiple Explainable AI (XAI) methods—SHAP, LIME, and DALEX—with various AI models to enhance both anomaly detection and interpretability. By fusing top features identified by these XAI methods across six diverse AI models (Decision Trees, Random Forests, Deep Neural Networks, K-Nearest Neighbors, Support Vector Machines, and AdaBoost), the framework creates a robust and comprehensive set of features critical for detecting anomalies. These  feature sets, produced by our feature ensemble framework, are evaluated using independent classifiers (CatBoost, Logistic Regression, and LightGBM) to ensure unbiased performance. We evaluated our feature  ensemble approach on two popular
autonomous driving datasets (VeReMi and Sensor) datasets.
Our feature ensemble technique demonstrates improved accuracy, robustness, and transparency of AI models, contributing to safer and more trustworthy autonomous driving systems.}





\maketitle

\input{Introduction}

\input{Background}

\input{Framework}

\input{Numerical_Simulations}

\input{Discussion}

\input{Conclusion}

\subsection*{Abbreviations}
\vspace{-1cm}

\nomenclature{ADS}{Anomaly Detection System}

\nomenclature{AI}{Artificial Intelligence}

\nomenclature{AV}{Autonomous Vehicle}

\nomenclature{CatBoost}{Categorical Boosting}

\nomenclature{CNN}{Convolutional Neural Network}

\nomenclature{DALEX}{Descriptive Analytics for Learning and EXplanation}

\nomenclature{DNN}{Deep Neural Network}

\nomenclature{DoS}{Denial of Service}

\nomenclature{DT}{Decision Tree}

\nomenclature{FDI}{False Data Injection}

\nomenclature{HDAD}{Hybrid Deep Anomaly Detection}

\nomenclature{KNN}{K-Nearest Neighbors}

\nomenclature{LGBM}{Light Gradient Boosting Machine}

\nomenclature{LIME}{Local Interpretable Model-Agnostic Explanations}

\nomenclature{LR}{Logistic Regression}

\nomenclature{LSTM}{Long Short-Term Memory}

\nomenclature{M-CNN}{Modified Convolutional Neural Network}

\nomenclature{RF}{Random Forest}

\nomenclature{SAMME.R}{Stagewise Additive Modeling using a Multiclass Exponential Loss Function, Real version}

\nomenclature{SHAP}{SHapley Additive exPlanations}

\nomenclature{SVM}{Support Vector Machine}

\nomenclature{VANET}{Vehicular Ad-hoc Networks}

\nomenclature{VeReMi}{Vehicular Reference Misbehavior}

\renewcommand{\nomname}{} \printnomenclature

\section*{Declarations}

\subsection*{Ethical Approval and Consent to participate}

Not applicable.

\subsection*{Consent for publication}

The authors give permission for this work to be published in the Journal of Transportation Security, and other publications produced by the  Journal of Transportation Security, in print and online.

\subsection*{Availability of supporting data}

The datasets generated and/or analysed during the current study are available in the GitHub repository, \url{https://github.com/Nazat28/XAI-based-Feature-Ensemble-for-Enhanced-Anomaly-Detection-in-Autonomous-Driving-Systems}.

\subsection*{Competing interests/Authors' contributions}

The authors declare that they have no competing interests.

\subsection*{Funding}

This work is supported in part by AnalytixIN, Enhanced Mentoring Program with Opportunities for Ways to Excel in Research (EMPOWER), and 1st Year Research Immersion Program grants from the office of the Vice Chancellor for Research at
Indiana University-Purdue University Indianapolis.

\subsection*{Author contribution}

Conceptualization: S.N. and M.A.; Data curation: S.N.; Formal analysis: S.N.; Funding acquisition: M.A.; Investigation: S.N.; Methodology: S.N. and M.A.; Project administration: M.A.; Resources: M.A.; Software: S.N.; Supervision: M.A.; Validation: S.N.; Visualization: S.N.; Writing – original draft: S.N.; Writing - review \& editing: S.N. and M.A.

\subsection*{Acknowledgements}

Not applicable.




\noindent

\bigskip





\bibliography{sn-bibliography}

\begin{appendices}

\section{Hyperparameters of AI Models}\label{app:hyperparam}


\textbf{(1) Decision Tree (DT):} Decision tree classifier was our first AI model that we experimented on the datasets. The best hyperparameter choice for this model was when the criterion was set to “gini” for measuring impurity and the max depth of the tree was 50 which means the number of nodes from root node to the last leaf node was 50. The minimum number of samples required to be present in a leaf node was 4 ($min\_samples\_leaf = 4$) and the minimum number of samples required to split a node into two child nodes was 2 ($min\_samples\_split = 2$). To have reproducibility and testing on same data we set $random\_state$ to 100 and the rest of the hyperparameters were set as default.

\textbf{(2) Random Forest (RF):} We next implemented RF classifier where $max\_depth$ was set to 50. The number of estimators (number of decision trees) was set to 100, $min\_samples\_leaf$ was set to 1, $min\_samples\_split$ was same as that of DT and the rest of the hyperparameters were used as provided by default.

\textbf{(3) Deep Neural Network (DNN):} We next show the best values for the DNN classifier. We set the dropout value to 0.1, added 1 hidden layer with a size of 16 neurons with rectified linear unit (ReLU) as the activation function. Next we set optimization algorithm as `ADAM' and the loss function was set to “binary\_crossentropy”. The epochs were set to 5 with batch size of 100 to train the DNN model. We set the rest of the hyperparameters as  given by default configuration.

\textbf{(4) K-nearest Neighbour (KNN)}: The parameters we set for KNN were as follows: the number of neighbors was set to 5 ($n\_neighbors=5$), the leaf size was set to 30 to speed up the algorithm,  the distance metric was set to “minkowski” to compute distance between neighbors, and the search algorithm for this model was “auto”. All other hyperparameters were used as set by default.

\textbf{(5) Support Vector Machine (SVM)}: For SVM AI classifier, we set the regularization parameter to 1 (C=1). The kernel function was set to radial basis function (RBF). Kernel coefficient to control decision boundary was set to “auto”. All other hyperparameters were set to default.

\textbf{(6) Adaptive Boosting (AdaBoost)}: Finally, the main hyperparameters we used for AdaBoost are as follows: “estimator” was set to `DecisionTreeClassifier' with a maximum depth of 50. The number of estimators was 200 and the “learning\_rate” was 1. The boosting algorithm was set to “SAMME.R” to converge faster with lower test error.
\vspace*{-8mm}

\end{appendices}


\end{document}

%% file: Introduction.tex
\section{Introduction}



The rapid advancement of autonomous vehicle (AV) technology has introduced significant challenges in transportation security. As AVs become more prevalent, ensuring their safety and reliability is paramount~\cite{Bagloee2016-yn}. Artificial Intelligence (AI) models have shown promise in detecting anomalies in the behavior of AVs~\cite{alqahtani2024machine}, but their black-box nature poses considerable obstacles to understanding and trusting their decision-making processes. This lack of interpretability is particularly concerning in the safety-critical domain of autonomous driving, where explainable decisions are crucial for public safety, user trust, and regulatory compliance~\cite{nazat2024evaluating}.

Current anomaly detection systems for AVs often rely on single AI models~\cite{han2022ads} or individual explainable AI (XAI) methods~\cite{nazat2024xai}. While these approaches have demonstrated promising results, they frequently fall short in capturing the full complexity of anomaly detection and providing robust and reliable explanations~\cite{al2019intrusion}. The key challenges in this context include:

\begin{itemize}
    \item \textbf{Incomplete Feature Importance Assessment}: Individual XAI methods often provide limited insights into feature importance, failing to capture the comprehensive set of factors influencing anomaly detection model's decisions~\cite{tritscher2023feature}.
    
    \item \textbf{Lack of Consensus Among XAI Methods}: Different XAI methods can yield conflicting interpretations, making it difficult to derive a consistent understanding of anomaly detection model's behavior~\cite{hassija2024interpreting}.
    
    \item \textbf{Insufficient Utilization of Multiple AI Models}: Relying on a single AI model limits the robustness of anomaly detection~\cite{nazat2024evaluating}, as different models may excel in different aspects of data interpretation.
    
    \item \textbf{Challenges in Feature Selection Optimization}: Effective anomaly detection requires identifying the most relevant features, a process that can be hindered by the limitations of using single XAI method~\cite{nazat2024xai}.
\end{itemize}

To help address these issues, this paper proposes a novel XAI-based feature ensemble framework that integrates multiple XAI methods (SHAP~\cite{shap2024}, LIME~\cite{lime2024}, and DALEX~\cite{dalex2024}) with various AI models to enhance anomaly detection in autonomous driving systems. Our approach  combines insights from different XAI methods to provide a more representative set of features that can better explain decision-making of anomaly detection models for AVs.

\textbf{Overview of Our Feature Ensemble Framework:}
Our framework operates as follows.
We use a diverse set of AI models, including Decision Trees (DT), Random Forests (RF), Deep Neural Networks (DNN), K-Nearest Neighbors (KNN), Support Vector Machines (SVM), and Adaptive Boosting (AdaBoost), to build anomaly detection models using the input data from autonomous vehicles. We then apply XAI methods (here SHAP, LIME, and DALEX) to these models to extract top features, providing a multi-faceted view of feature importance.
The top features identified by each XAI method are combined using a frequency analysis technique to create a unified set of features, capturing the most critical aspects of the data.  

We evaluate our framework using two autonomous driving datasets, VeReMi~\cite{van2018veremi} and Sensor~\cite{9257492}, which represent different aspects of autonomous vehicle behavior. The VeReMi dataset focuses on vehicular positioning and speed in x, y, and z directions, while the Sensor dataset encompasses data from multiple sensors used in AVs. To validate the effectiveness of our feature ensemble approach, we employ three independent classifiers which are Categorical Boosting (CatBoost), Light Gradient Boosting Machine (LGBM), and Logistic Regression (LR). These classifiers were chosen for their diverse algorithmic approaches and proven effectiveness in various machine learning tasks~\cite{saheed2021ensemble}. In particular, we  feed the ensemble feature sets into these three independent classifiers to avoid bias in the decision-making of AI models used to generate these features.

Our evaluation results demonstrate that the proposed XAI-based feature ensemble approach consistently performs on par with, and in some cases outperforms, individual XAI methods across these independent classifiers. The key findings from our evaluation results include:

\begin{itemize}
    \item \textbf{Robust Feature Set for Anomaly Detection}: The fusion of features from multiple XAI methods provides a more robust set of indicators for anomaly detection. Our approach achieved accuracy rates of up to 82\% on the VeReMi dataset for binary class classification and 82\% on the Sensor dataset using the CatBoost classifier.
    
    \item \textbf{High Performance Across Classification Tasks}: Our framework maintains high performance across different classification tasks, including binary and multiclass classification. For multiclass classification on the VeReMi dataset, our approach achieved an F1-score of 0.80 across the three independent classifiers (CatBoost, LGBM, and Logistic Regression classifiers).
    
     \item \textbf{Consistency and Generalizability}: The performance of our feature ensemble approach is consistent across the three independent classifiers, demonstrating its robustness and generalizability. This consistency indicates that our methodology can be effectively applied to various AI models and datasets.
\end{itemize}

Given these merits of our work, this study  contributes to the development of more reliable, interpretable, and secure autonomous driving systems by bridging the gap between high-performance anomaly detection and identifying explanatory features using explainable AI. Our framework represents a significant step towards trustworthy AI in safety-critical autonomous vehicle applications, paving the way for future advancements in this critical field. By integrating multiple XAI methods and employing a comprehensive feature ensemble technique, we enhance the accuracy, interpretability, and robustness of anomaly detection in AVs, thereby contributing to the broader goal of safer and more reliable autonomous transportation systems.

\subsection{Summary of Contributions} 
The main contributions of this paper can be summarized as follows.

 \begin{itemize}
\item 

We propose a novel feature ensemble approach that combines SHAP, LIME, and DALEX XAI methods to enhance feature importance analysis, leveraging the strengths of each method for a more comprehensive understanding. The proposed approach is a frequency-based feature ensemble technique that creates a unified set of top features.

\item We apply our feature ensemble approach using six diverse AI models (DT, RF, DNN, KNN, SVM, AdaBoost) to gain insights into feature importance, identifying common important features in these models and enhancing the robustness of the findings.

\item We apply our feature ensemble approach  on two popular autonomous driving datasets (VeReMi and Sensor), contributing to the understanding of critical features in autonomous vehicle security and anomaly detection.

\item  We evaluate the effectiveness of the fused feature set using independent classifiers (CatBoost, LR, and LGBM), demonstrating performance improvements and practical utility in enhancing anomaly detection.

\item We release our source codes for the research community.
\footnote{The URL for our  source code is: \url{https://github.com/Nazat28/XAI-based-Feature-Ensemble-for-Enhanced-Anomaly-Detection-in-Autonomous-Driving-Systems}}


\end{itemize}

\section{Related Works}

\subsection{Anomaly detection in Autonomous Driving}

Numerous studies have investigated anomaly detection in autonomous driving systems~\cite{iasc.2022.020936,zekry2021anomaly,sym14071450}. For instance, in~\cite{iasc.2022.020936}, a modified convolutional neural network (M-CNN) was applied to onboard and external sensor measurements to detect instantaneous anomalies in an autonomous vehicle (AV). This approach focuses on identifying sudden changes or spikes in sensor data values or abrupt changes in GPS location coordinates. The study~\cite{zekry2021anomaly} employed a combination of convolutional neural networks (CNN) and Kalman filtering to detect abnormal behaviors in AVs. Meanwhile, the authors in~\cite{sym14071450} used long short-term memory (LSTM) networks to identify false data injection (FDI) attacks, ensuring the stable operation of AVs. Our work, however, emphasizes anomaly detection through the lens of explainable AI (XAI) and feature understanding. Several works have also explored anomaly detection in networks of vehicles~\cite{9815151,8684317,9210741,10.1145/3485832.3485883}. The study~\cite{9815151} proposed a hybrid deep anomaly detection (HDAD) framework, enabling AVs to detect malicious behavior based on shared sensor network data. Additionally, the research~\cite{8684317} utilized time-series anomaly detection techniques to identify cyber attacks or faulty sensors. The prior work~\cite{9210741} leveraged a CNN-based LSTM to classify signals from multiple sources as either anomalous or healthy in AVs. In contrast, our framework introduces a novel method to identify significant features of an AV which are taken into account to classify the AV as benign or anomalous using different well-known XAI techniques.

\subsection{Explanation using XAI}

In the AI domain, several studies have employed XAI methods to enhance the performance of AI models. Previous research~\cite{apicella2023strategies} utilized various XAI techniques, such as Saliency, Guided Backpropagation, and Integrated Gradients, to determine if these methods could improve model performance beyond merely providing explanations in the context of image classification. Another study~\cite{apicella2022xai} introduced an innovative XAI-based masking approach that uses integrated gradients explanations to enhance image classification systems. Additionally,~\cite{apicella2022toward} examined the explanations generated by XAI methods for an EEG emotion classification system.  However, none of these studies provided the top robust significant features which contributes to the anomaly detection (and classification) of AVs.  

In the autonomous driving domain, the work \cite{madhav2022explainable} provided a broader overview of XAI in AVs, focusing on intelligent transportation systems. While they explored visual explanatory methods and an intrusion detection classifier, their approach lacks the methodological specificity of our feature ensemble framework. Their study offers a general comparison of XAI applications but does not integrate multiple XAI methods or provide comprehensive evaluations across diverse AI models and independent classifiers, as we do here in our current work. Additionally, they addressed transparency in decision-making but omit the systematic feature importance analysis and anomaly detection enhancements central to our research, highlighting the distinct contribution of our approach to XAI in AV security. Both our study and \cite{atakishiyev2024explainable} focus on the importance of XAI in AVs, aiming to enhance transparency, trust, and reliability. While both emphasize making AI decisions understandable for regulatory and social acceptance, our work introduces a specific feature ensemble framework using multiple XAI methods (SHAP, LIME, DALEX) for anomaly detection. In contrast, their study offers a broader overview of XAI approaches in AVs. Together, these studies highlight complementary approaches to improving transparency and trust in autonomous driving systems. Again, both our study and \cite{kuznietsov2024explainable} emphasize the role of XAI in improving the safety and trust of autonomous driving systems by making AI decisions more transparent. While our research introduces a feature ensemble framework for anomaly detection using multiple XAI methods, their work offers a systematic review of XAI techniques and presents the SafeX framework. However, these studies provide complementary approaches: ours with a concrete methodology, and theirs with a broader high-level conceptual framework, while both addressing the challenges of building safe autonomous vehicles.

\subsection{Contributions of This Work}
In our work, we consider the top features from three XAI methods from six AI models and fuse them using frequency analysis in order to get a more stable and robust set of features that are responsible for the classification of AVs. Afterwards, we feed these feature sets to three independent classifiers to make sure there is no bias in the performance. This multi-layered approach offers a more holistic and dependable method for identifying critical features in anomaly detection for AVs, addressing the limitations of previous single-AI or single-XAI approaches.

Our framework leverages multiple explainable AI (XAI) methods across various black-box AI models. Specifically, we employ three XAI methods (SHAP, LIME, and DALEX) in conjunction with six AI models (Decision Trees, Random Forests, Deep Neural Networks, K-Nearest Neighbors, Support Vector Machines, and AdaBoost) to identify important features. We then use a frequency-based fusion approach to consolidate these features into a unified set. The effectiveness of this fused feature set is evaluated using three independent classifiers (CatBoost, LightGBM, and Logistic Regression) and compared against the performance of individual XAI methods on two distinct datasets.

%% file: Background.tex
\section{The Problem Statement}\label{sec:background}



We now outline the key challenges in anomaly detection for autonomous driving systems, including the limitations of black-box AI models, and present our proposed framework that integrates multiple XAI methods and AI models to enhance accuracy and interpretability of these systems.

\subsection{Challenges in Anomaly Detection for Autonomous Vehicles}


Ensuring the safety and reliability of AVs has become paramount. AI models have shown promise in detecting anomalies in AV behavior~\cite{aminanto2016deep}, but their black-box nature poses considerable obstacles to understanding and trusting their decision-making processes. This lack of interpretability is particularly concerning in the safety-critical domain of autonomous driving, where explainable decisions are crucial for public safety, user trust, and regulatory compliance~\cite{dixit2022anomaly}. 

Current anomaly detection systems for AVs often rely on single AI models or individual XAI methods. While these approaches have demonstrated some success, they  have several limitations.
First, a single XAI method may fail to identify all critical features necessary for effective anomaly detection. Second, relying on a single AI model for anomaly detection limits the ability to leverage the diverse strengths of various machine learning algorithms. Third, determining the optimal set of features for anomaly detection is a complex task that requires balancing model performance with interpretability and computational efficiency. Traditional methods may struggle to achieve this balance.

To address these challenges, a more robust and comprehensive approach is needed—one that integrates multiple AI models and XAI methods to provide a holistic understanding of feature importance and anomaly detection. This approach should aim to enhance the interpretability, reliability, and overall effectiveness of anomaly detection systems for autonomous vehicles.

\subsection{Main Objectives for Enhanced Anomaly Detection in Autonomous Vehicles}

To address these challenges, there is an imperative need for a novel framework that integrates multiple Explainable AI (XAI) methods and AI models to enhance the accuracy and interpretability of anomaly detection in autonomous driving systems. Such a framework should be designed with the following key objectives:

\textbf{a) Synthesize Insights from Various XAI Techniques:} By employing a range of XAI methods, such as SHAP, LIME, and DALEX, the framework can provide a holistic and detailed understanding of feature importance. Each XAI technique offers unique insights and strengths, and their combined application can uncover critical features that may be overlooked when using a single method. This comprehensive synthesis ensures a deeper and more accurate analysis of the features influencing AV's behavior.

\textbf{b) Develop a Fusion Methodology:} The framework must incorporate a robust fusion methodology to reconcile potentially conflicting feature rankings generated by different XAI methods. This process involves performing a frequency analysis to determine the most consistently important features across various methods and models. By integrating these diverse insights, the fusion methodology will create a unified and reliable feature ranking that enhances the effectiveness of anomaly detection.

\textbf{c) Leverage the Strengths of Multiple AI Models:} Different AI models excel in various aspects of data analysis and anomaly detection. The framework should integrate multiple well-known AI models, such as Decision Trees, Random Forests, K-Nearest Neighbors, Support Vector Machines, Deep Neural Networks, and AdaBoost, to harness their complementary strengths. This feature ensemble approach will improve the overall performance of the anomaly detection system by leveraging diverse capabilities of these models.

\textbf{d) Optimize Feature Selection:} The framework should optimize feature selection to balance performance,
interpretability, and computational efficiency. This involves identifying the most relevant and impactful features while ensuring that the resulting models remain interpretable and computationally feasible. Effective feature selection will enhance the anomaly detection system’s accuracy without compromising its  usability.

By addressing these critical objectives, the proposed framework aims to significantly advance the state-of-the-art secure anomaly detection models for autonomous vehicles. It will help in identifying and understanding anomalous AV behavior, ultimately contributing to safer and more trustworthy autonomous driving systems. 

%% file: Framework.tex
\section{Framework}\label{sec:framwork}


\subsection{Adversary and Defense Models}

\textbf{Adversary Model:} To rigorously test the robustness of our XAI-enhanced anomaly detection framework, we describe the adversary model featuring a sophisticated threat actor targeting Vehicular Ad-hoc Networks (VANETs). We focus on data falsification attack scenario, where sensor readings (e.g., position and speed in x, y, and z directions for the VeReMi dataset) are subtly altered by the attacker. The adversary has limited knowledge of the defense model's architecture,  the capacity to introduce or compromise AVs, however it can generate and inject realistic falsified data. This model  challenges our defense system to validate the efficacy and robustness of our XAI-based feature ensemble  approach in identifying subtle and sophisticated anomalies.

\textbf{Defense Model:} In our defense model for autonomous vehicles (AVs), we equip each AV with different  distinct sensors to gather data during inter-vehicle communications, mirroring previous research methodologies~\cite{9257492}. This sensor data is concatenated to form a comprehensive input for our anomaly detection classifiers. Our innovative approach involves applying multiple XAI methods—SHAP, LIME, and DALEX—to this data across various AI models. We then employ a feature ensemble technique, consolidating the top features identified by each XAI method through frequency analysis. This fused feature set is used to train independent classifiers, enhancing the accuracy and interpretability of anomaly detection in vehicular networks. This methodology aims to pinpoint the most critical sensors and data points for distinguishing between benign and anomalous AV behavior.

\begin{figure*}[t]
   \centering
\includegraphics[width=0.8\linewidth]{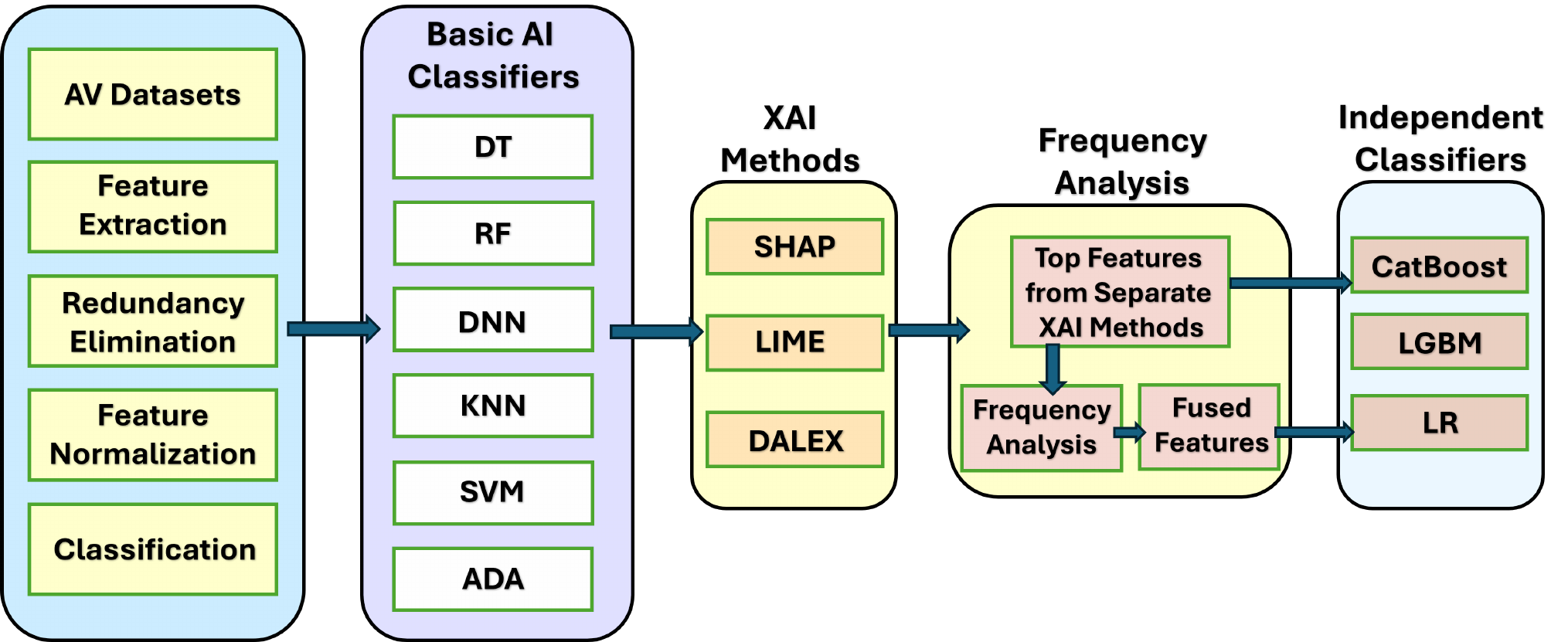}
   \caption{An overview of different components of our XAI-based feature ensemble framework for enhancing anomaly detection of autonomous vehicles.}  
   \label{fig:Overview}
   \vspace{-2mm}
\end{figure*}

\subsection{An End-to-End  Feature Ensemble Pipeline for Autonomous Driving Systems}\label{sec:framework_comp}


The primary objective of this research is to develop a pipeline that enhances our understanding of the main features of autonomous vehicles (AVs) and the decision-making processes of anomaly detection AI models in autonomous driving systems. To accomplish this, we propose a comprehensive end-to-end framework that leverages the synergistic power of multiple explainable AI methods. Our approach uniquely combines SHAP, LIME, and DALEX to analyze feature importance across diverse AI models, including decision trees, random forests, k-nearest neighbors, support vector machines, and adaptive boosting. By fusing the outputs of these XAI methods through frequency analysis, we derive a consolidated set of top features for each dataset. This novel feature ensemble methodology aims to provide a better understanding of critical factors in anomaly detection for autonomous driving systems. 

%
%
%
The different components of our pipeline (shown in Figure~\ref{fig:Overview}) are explained in details below.


\textbf{Loading Autonomous Driving Dataset:} In this study, we employ two distinct datasets for the anomaly classification of autonomous vehicles (AVs). The first dataset is the Vehicular Reference Misbehavior (VeReMi) dataset, designed to analyze misbehavior detection mechanisms in vehicular ad hoc networks (VANETs). Generated from a simulation environment, it provides message logs of on-board units (OBU) and labeled ground truth data~\cite{vanderheijden2018veremi}. The second dataset is based on Sensor data~\cite{9257492}, aimed at monitoring unusual activity from an AV using data gathered from ten distinct sensors on the vehicle. It is important to note that, when selecting the Sensor data for each AV, we adhered to the data ranges specified in previous works~\cite{inproceedings}, \cite{5604050}.

\begin{table*}[th]
\centering
\caption{The main sensors used for anomaly detection task of each AV for the Sensor dataset used in this work.}
\vspace{-1mm}
\resizebox{0.99\linewidth}{!}{
\vspace{-1mm}
\begin{tabular}{|l|l|l|} 
\hline
\textbf{Sensor Name}    & \textbf{Normal Data Range} & \textbf{Description}                                                                                             \\ 
\hline
Formality      & 1bit – 10bit      & Checks every message if it is maintaining correct formality                                               \\ 
\hline
Location       & 0/1               & Checks if the message reached the destined location                                                       \\ 
\hline
Frequency      & 1Hz – 10Hz        & Checks the interval time of messages                                                                      \\ 
\hline
Speed          & 50mph – 90mph     & Checking if the AV is in the speed limit (Highway)                                                        \\ 
\hline
Correlation    & 0/1               & Checks if several messages adhere to defined specification  \\ 
\hline
Lane Alignment & 1-3             & Checks if the AV is in the correct lane                                                                          \\ 
\hline
Headway Time   & 0.3s - 0.95s      & Checks if the AV maintains the headway time range                                                         \\ 
\hline
Protocol       & 1-10000           & Checks for the correct order of communication messages                                                                          \\ 
\hline
Plausibility   & 50\% - 200\%      & Checks if the data is plausible (relative difference between sizes of two consecutive payloads).                                                     \\ 
\hline
Consistency    & 0/1               & Checks if all the parts of the AV are delivering consistent information about an incident                 \\
\hline
\end{tabular}
}
\vspace{-2mm}
\label{tbl:ten_sensors_and_values_per_AV}
\end{table*}

\textbf{Feature Extraction:} After loading the datasets, we select the essential features from each dataset to build our AI models. Feature extraction is crucial in autonomous driving to mitigate adversarial attacks~\cite{limbasiya2022systematic}. By identifying features indicative of attacks, the overall attack surface can be reduced. For the VeReMi dataset, we selected the top six most important features, following the methodology outlined in prior work~\cite{mankodiya2021xai}. Conversely, for the Sensor dataset, we utilized all ten features to train our models, as inspired by previous studies~\cite{inproceedings, 5604050}.

\textbf{Redundancy Elimination:} The next step in our framework involves removing redundancy from our datasets. Redundancy is common in large datasets and can adversely affect an AI model's performance. For the VeReMi dataset, we removed identical rows and those containing empty features. In contrast, for the Sensor dataset, we did not find any redundancy.

\textbf{Classification Problem:} Although both datasets (VeReMi and Sensor) contain multiple types of anomalies, we generalize all anomaly classes as ``anomalous,'' resulting in a binary classification with two classes: benign ("0") and anomalous ("1"). This approach aligns with our objective of detecting anomalies irrespective of their specific source and facilitates the implementation of a general AI-based framework for anomaly detection in AVs. Additionally, we consider multi-class classification for the VeReMi dataset, with the evaluation results provided in Section~\ref{evaluation_results}.

\textbf{Data Balancing:} Our next step involve addressing the imbalance in the VeReMi dataset, which initially contained a significant number of normal data points compared to anomalous ones. To achieve a balanced dataset, we apply the random under-sampling method as described in prior work~\cite{undersampling2021}. This method involves removing data from the majority class (normal) that are considered less informative to the model, thereby creating a balanced distribution. Random under-sampling is widely recognized for its efficiency in balancing datasets due to its straightforward implementation and low complexity~\cite{liu2020dealing}. In contrast, the Sensor dataset did not require such balancing measures as its inherent setup was predominantly balanced.

\textbf{Feature Normalization:} After performing basic feature extraction and data balancing, we proceed with feature normalization on our two autonomous driving datasets. This step, crucial in the data preprocessing phase, ensures that the features are scaled to a common range or distribution, significantly impacting the effectiveness of AI algorithms. The purpose of normalization is to prevent any single feature from dominating the analysis. We achieve this by using standard scalar feature scaling (see~\cite{scikit-learn} for an example), which transforms the feature values to have a mean of 0 and a standard deviation of 1.


\textbf{Black-box AI Models and Related Evaluation:} After finishing the previous stages, we proceed to train the anomaly detection AI models utilizing 70\% of the data. For every model, we optimize its hyperparameters to achieve the optimal outcome. After the completion of the training, we assess the models using a 30\% portion of the data that has not been previously examined during training. Our framework's subsequent step involves assessing the outcomes of black-box AI algorithms. We calculate several assessment metrics for each output of the AI model, including accuracy (Acc), precision (Prec), recall (Rec), and F1-score (F1)~\cite{scikit-learn}. 

\textbf{XAI Global Explanation:} It is important to note that the AI models we create and assess in the preceding stage are classified as black-box AI models. Hence, it is imperative to furnish elucidations regarding these models and their corresponding features and AV classification (benign or anomalous). Therefore, the subsequent phase in our architecture is the XAI step. In this study, we specifically focus on the overall interpretation of our three XAI models. We utilize the SHAP global summary plot, which assumes that the global relevance of each feature is represented by the mean absolute value of that feature across all input samples. Since LIME is specifically designed for providing local explanations of samples, we have devised Algorithm~\ref{alg:lime_feature_importance} to utilize LIME for  incorporating global explanations.

\begin{algorithm*}[t]
\caption{Algorithm for Generating Top Features Using LIME}
\label{alg:lime_feature_importance}
\noindent \textbf{Input:} Dataset, and Trained model

\textbf{Output:} Sorted list of features along with their average importance scores

Create an explainer object using LIME's LimeTabularExplainer

Initialize feature lists and dictionaries to store feature importance

\textbf{For} {each instance in the dataset (e.g., 50,000 samples)}
    
    \hspace*{3mm} Use LIME to generate instance's explanation based on the model's prediction
    
    \hspace*{3mm} Extract the importance scores for all the features from the explanation
    
    \hspace*{3mm} \textbf{For} {each feature, importance in explanation}
    
    \hspace*{3mm}  Accumulate absolute importance scores for each feature across all samples\newline
    \hspace*{3mm} \textbf{End}\newline
\textbf{End}

Normalize the accumulated importance scores by dividing them by the total number of samples to obtain average importance scores for each feature

Sort the features based on their average importance scores in descending order

Present the sorted list of features along with their average importance scores

\textbf{Return} Sorted list of features and their average importance scores
\end{algorithm*}

For DALEX, we utilize its global explainability feature, which provides an explanation of the significance of each feature considered by the AI models in classifying whether an AV is benign or abnormal.

 \textbf{Novel Feature Fusion Framework:} Recall that appropriate feature selection plays a significant role in the performance of AI models. In this study, we propose XAI-based novel methods for feature extraction aimed at enhancing the performance of AI models in anomaly detection within autonomous driving systems. Our approach integrates three XAI methods—SHAP, LIME, and DALEX—applied across five AI models (Decision Trees (DT), Random Forests (RF), K-Nearest Neighbors (KNN), Support Vector Machines (SVM), and AdaBoost) using two distinct datasets (VeReMi and Sensor).

\textbf{(a) XAI-based Feature Ranking:} The XAI-based feature ranking methodology is structured as follows:

\begin{itemize}
    \item  Identify the primary columns/features for each dataset.
    \item  Utilize SHAP, LIME, and DALEX to generate global explanations for each of the six AI models.
    \item  Use the explanation values to evaluate each feature's contribution to the model's performance.
    \item  Rank features based on their average importance values across all separate six models and three XAI methods.
    \item  Extract the top-k ranked features to input into the independent AI models (CatBoost, LGBM, and LR) while varying k (number of top features) to assess the impact on anomaly detection performance. 
\end{itemize}

In other words, we start by identifying the primary features for each dataset. Then, we use three Explainable AI (XAI) methods—SHAP, LIME, and DALEX—to generate global explanations for six AI models (Decision Trees, Random Forests, K-Nearest Neighbors, Support Vector Machines, Deep Neural Networks, and AdaBoost). These explanations allowed us to assess each feature's contribution to the model's performance. We then ranked the features based on their average importance values across all models and XAI methods. Finally, we extract the top-k ranked features to input into the AI models, while varying k to evaluate the impact of feature ensemble on anomaly detection performance.

\textbf{(b) XAI-based Feature Ensemble Ranking:}
Our feature ensemble ranking method follows these steps:

\begin{itemize}
    \item Train six AI models (DT, RF, KNN, SVM, DNN, AdaBoost) on the dataset.
    \item Apply three XAI methods (SHAP, LIME, and DALEX) to each trained model to generate sets of top features.
    \item Analyze the frequency of each feature appearing in the top rankings across all XAI methods and models, considering the ranking position.
    \item Calculate the final importance of each feature using a weighted scoring system:
    \begin{itemize}
        \item A feature receives 3 points for each first-place ranking.
        \item It receives 2 points for each second-place ranking.
        \item It receives 1 point for each third-place ranking.
        \item The formula used for final feature importance is: 
        Final Feature Importance Score = $A \times 3 + B \times 2 + C \times 1$, where:
        \begin{itemize}
            \item $A$ = number of times the feature ranked first.
            \item $B$ = number of times the feature ranked second.
            \item $C$ = number of times the feature ranked third.
        \end{itemize}
    \end{itemize}
    \item Rank features based on their above importance scores.
    \item Use the final ranked list of features as input for independent classifiers (here, CatBoost, LR, and LGBM) to evaluate the effectiveness of the feature ensemble method in improving anomaly detection.
\end{itemize}

To explain further, we implement an XAI-based feature ensemble ranking method to enhance anomaly detection performance in autonomous driving systems. Initially, we train six AI models DT, RF, DNN, KNN, SVM, and AdaBoost on our dataset. Subsequently, we apply three XAI methods—SHAP, LIME, and DALEX—to each trained model, generating distinct sets of top features from each method. We then perform a feature frequency analysis to determine the prevalence and ranking positions of features across all XAI methods and models. Then, to quantify feature importance, we employ a weighted scoring system, awarding 3 points for first-place rankings, 2 points for second-place, and 1 point for third-place for each XAI method. We want to underscore that we take top-$k$ features from each of our dataset ($k$=4, 3, and 2 for VeReMi and 5 for Sensor). After having the frequency-based features from our three XAI methods, we incorporate the same frequency analysis to fuse these three sets of features to one. This approach results in a comprehensive final ranking of features based on their computed importance scores. Finally, we utilize this ranked list to feed top-k features into independent classifiers (CatBoost, Logistic Regression, and LightGBM), assessing the efficacy of our feature ensemble method in improving anomaly detection. This XAI-based feature ensemble method integrates insights from multiple XAI techniques and models, providing a holistic approach to feature understanding in the context of autonomous driving.

\subsection{Lists of Top Features for Anomaly Detection in the Used Autonomous Driving Datasets}

We now present the complete lists of primary characteristics (features) used in constructing our anomaly detection AI models, along with their explanations, for the VeReMi and Sensor datasets utilized in our framework. Tables~\ref{tbl:ten_sensors_and_values_per_AV} and~\ref{tbl:veremi_data_features} describe each feature in the Sensor and VeReMi datasets, respectively. Our XAI framework will be employed to derive key insights from these features for detecting anomalous AV behavior in these datasets (as will be shown in Section~\ref{sec:evaluation}).

\subsection{Illustrations of Global Explanations by the Three XAI methods}

We now present three illustrative instances of XAI global explanations drawn from the three XAI models considered in this work. We utilized DT to illustrate the XAI approaches we are using in this work, as an example of how we see the features and their contributions to our VeReMi dataset.

\begin{figure}[ht]

\centering

\begin{subfigure}[b]{0.49\textwidth}
\centering
\includegraphics[width=\linewidth]{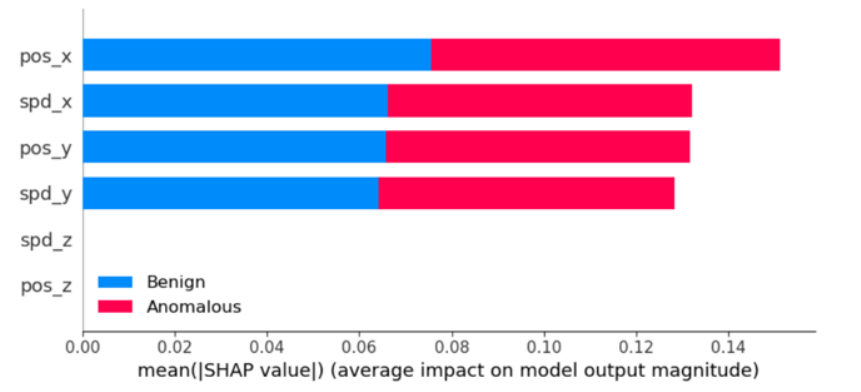}
\caption{SHAP}
\label{SHAP}
\end{subfigure}
~
\begin{subfigure}[b]{0.45\textwidth}
\centering
\includegraphics[width=\linewidth]{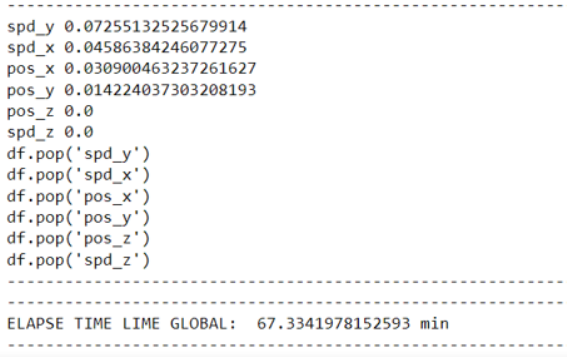}
\caption{LIME}
\label{LIME}
\end{subfigure}

\begin{subfigure}[b]{0.55\textwidth}
\centering
\includegraphics[width=\linewidth]{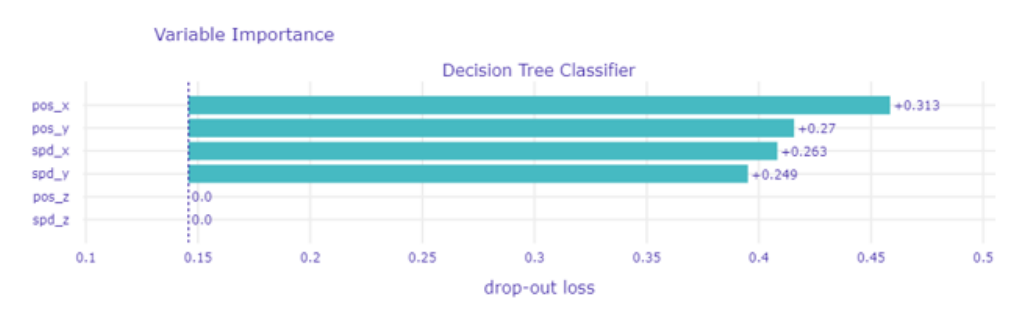}
\caption{DALEX}
\label{DALEX}
\end{subfigure}

\caption{An example of the three XAI methods (SHAP, LIME, and DALEX) eliciting the top features contributing to the anomaly detection of AV for VeReMi dataset.}
\label{Three_XAI_methods}
\end{figure}

\textbf{Global Explanations using SHAP:} Figure~\ref{SHAP} illustrates the top four features in the VeReMi dataset that had the greatest impact on anomaly identification for AVs in the x, y, and z directions for both position and speed in SHAP. Furthermore, for SHAP, the figure demonstrates that in this particular scenario, the spd\_z and pos\_z features do not contribute in any way to the detection of anomalies.

\textbf{Global Explanations using LIME:} Now in Figure~\ref{LIME}, it is depicted that spd\_y is the top ranked feature in case of LIME. On the other hand, spd\_x, pos\_x, and pos\_y comes in second, third and fourth position respectively in the anomaly detection contribution for VeReMi dataset. Again, for this case, pos\_z and spd\_z did not have any impact.

\textbf{Global Explanations using DALEX:}
In Figure~\ref{DALEX}, the top ranked feature for DALEX is pos\_x. pos\_y, spd\_x and spd\_y come subsequently to contribute to the anomaly detection. Similarly to both the previous cases, pos\_z and spd\_z did not influence to the anomaly detection of AVs at all.

Having explained the main components of our XAI-based feature ensemble framework for enhancing anomaly detection for autonomous driving systems, we next provide a thorough evaluation of our proposed pipeline.

\begin{table*}[t]
\centering
\vspace{2mm}
\caption{Description of main features of VeReMi dataset.}
\begin{tabular}{cl}
Column       & Description                             \\ \hline
pos\_x       & The x-coordinate of the vehicle position\\
pos\_y       & The y-coordinate of the vehicle position \\
pos\_z       & The z-coordinate of the vehicle position \\
spd\_x       & Speed of the vehicle in the x-direction         \\
spd\_y       & Speed of the vehicle in the y-direction         \\
spd\_z       & Speed of the vehicle in the z-direction       
\end{tabular}
\label{tbl:veremi_data_features}
\vspace{-2mm}
\end{table*}

%% file: Numerical_Simulations.tex

\section{Foundations of Evaluation}\label{sec:evaluation}





We next present our detailed evaluation setup. Our evaluation aims to answer the following questions: 

\begin{itemize}

    \item What is the overall performance of different AI models when applied to the autonomous driving datasets (VeReMi and Sensor)?

    \item How effective are the top features identified by the three XAI methods (SHAP, LIME, and DALEX) in enhancing anomaly detection model's performance?

    \item How does the frequency-based feature ensemble technique impact the performance of independent AI classifiers (CatBoost, LR, and LGBM)?




\end{itemize}


To address these questions, we conducted experiments using three XAI methods (SHAP, LIME, and DALEX) to extract top features from the VeReMi and Sensor datasets, employing six AI models (DT, RF, DNN, KNN, SVM, AdaBoost). After obtaining the top features from each XAI method, we performed frequency analysis to generate a unified set of top features for each dataset. This consolidated feature set was then fed into three independent AI models (CatBoost, LR, and LGBM) to evaluate potential improvements in performance. We start by explaining the two autonomous driving datasets used in our study and the experimental setup designed to answer these questions.


\subsection{Dataset Description}

\textbf{VeReMi dataset~\cite{vanderheijden2018veremi}:} 
The VeReMi dataset serves as a significant resource for anomaly detection research in autonomous driving systems. It encompasses a range of attack scenarios, including Denial of Service (DoS), Sybil attacks, and message falsification within Vehicular Ad-hoc Networks (VANETs). The dataset's strength lies in its provision of real-world data, complete with sensor readings and corresponding ground truth labels for each attack type. Comprising 225 distinct scenarios and five attacker classifications, VeReMi includes both autonomous vehicle (AV) message logs and attacker ground truth files~\cite{vanderheijden2018veremi}. These components enable researchers to analyze attacker characteristics in depth. The dataset's comprehensive nature has established it as a benchmark in autonomous driving security research.

In our analysis, we focused on extracting features that best characterize AV behavior. Through a process of feature selection, we narrowed our focus to six key variables: pos\_x, pos\_y, pos\_z, spd\_x, spd\_y, and spd\_z. These parameters represent the three-dimensional position and velocity components of an AV, respectively. We determined that other available features did not contribute substantial additional information to our analysis. This targeted approach to feature selection allows for a more focused examination of AV dynamics-related features while potentially reducing computational complexity in subsequent analyses.

\textbf{Sensor dataset:} In our research, we also utilized the Sensor dataset to evaluate our framework, following the methodology outlined in the work~\cite{5604050}. This dataset comprises ten features, each simulating a distinct sensor presumed to be present in autonomous vehicles (AVs). These sensors include formality, location, speed, frequency, correlation, lane alignment, headway time, protocol, plausibility, and consistency. Each sensor serves a specific function in monitoring AV behavior, described as follows.

\begin{itemize}
    \item The formality sensor verifies message size and header integrity.
    \item Location sensor confirms message delivery to the intended recipient.
    \item Speed sensor ensures data remains within prescribed speed limits.
    \item Frequency sensor monitors message timing behavior.
    \item Correlation sensor checks for adherence to defined specifications across different messages.
    \item Lane alignment sensor verifies the AV's position within its designated lane.
    \item Headway time sensor assesses maintenance of appropriate following distances.
    \item Protocol sensor validates the correct sequencing of messages.
    \item Plausibility sensor evaluates message plausibility by examining relative size differences between consecutive messages.
    \item Consistency sensor ensures data coherence across various sources.
\end{itemize}

These sensors collectively contribute to determining the operational mode (normal or malicious) of each AV. The Sensor dataset provides normal data ranges for each feature, allowing for the identification of anomalous behavior in AVs that deviate from these established norms. A comprehensive description of each sensor (feature) and its associated values for individual AVs is presented in Table~\ref{tbl:ten_sensors_and_values_per_AV}. This detailed breakdown facilitates the understanding of dataset's structure and specific parameters used to assess AV behavior.

\textbf{Summary and Statistics of the Datasets:} Table~\ref{tbl:statistics_of_2_datasets} provides a detailed summary of the two datasets, including metrics such as dataset size, number of features, number of labels, and attack types. We partitioned each dataset into 70\% for training and 30\% for testing. Using this partitioning strategy, we developed six widely recognized AI classification models, described in detail below. 

\begin{table*}[h]
\centering
\caption{Statistics of both VeReMi and Sensor datasets.}
\begin{tabular}{|l|c|c|}
\hline
                  & \textbf{VeReMi dataset} & \textbf{Sensor dataset} \\ \hline
Number of Labels            & 2 and 6                & 2                \\ \hline
Number of Features          & 6                & 10               \\ \hline
Dataset Size  & 993,834           & 10,000            \\ \hline
Training Sample & 695,684           & 7,000            \\ \hline
Testing Sample  & 298,150           & 3,000            \\ \hline
Normal Samples No.       & 664,131           & 5,000            \\ \hline
Anomalous Samples No.           & 329,703           & 5,000 \\ \hline
\end{tabular}
\label{tbl:statistics_of_2_datasets}
\vspace{-4mm}
\end{table*}


\subsection{Experimental Setup}





\textbf{Coding Tools:} We used several open-source tools (based on Python) and various black-box AI models using well-known libraries like Keras~\cite{chollet2015keras} and Scikit-learn~\cite{scikit-learn} in order to create our  feature ensemble framework.

\textbf{XAI Methods:} Our framework incorporates three XAI methods: SHAP, LIME, and DALEX, detailed below.

\textbf{(a) SHAP~\cite{SHAP2022}:} SHAP explains AI model predictions by evaluating feature importance. Rooted in game theory (Shapley values), it assesses the contribution of each feature to the model's classification decisions.

\textbf{(b) LIME~\cite{LIME2020}:} LIME is an XAI method that provides local explanations for AI model predictions, making each prediction comprehensible on an individual basis. For each instance, LIME generates a local surrogate model that approximates the global AI model, detailing the features that influenced the classifier's decision for that specific instance.


\textbf{(c) DALEX~\cite{biecek2021explanatory}:} DALEX is a model-agnostic toolkit designed to interpret machine learning models by creating an explainer object that integrates the model, data, and response variables. It features tools for assessing feature importance, visualizing feature relationships through Partial Dependence Plots (PDP) and Accumulated Local Effects (ALE), identifying model weaknesses via residual diagnostics, and decomposing individual predictions with break down plots. By enhancing model interpretability and transparency, DALEX is crucial for building trust in AI systems across various algorithms.




\textbf{AI Models:} In our experimental evaluation, we employed six diverse AI models—Decision Tree (DT), Random Forest (RF), Deep Neural Network (DNN), k-Nearest Neighbor (KNN), Support Vector Machine (SVM), and Adaptive Boosting (AdaBoost)—to assess the efficacy of our feature ensemble framework. This selection spans traditional machine learning algorithms to more complex architectures, enabling us to examine the framework's performance across varying model complexities. These models were applied to two autonomous driving datasets to elucidate their black-box characteristics via our XAI framework. We carefully tuned hyperparameters for optimal performance and comparability,  as detailed in Appendix~\ref{app:hyperparam}.

\textbf{Metrics for Black-box AI:} To evaluate the outcomes of the black-box AI models and the efficiency of the proposed feature ensemble framework, we generated a standardized set of evaluation metrics to analyze each model's classifications. The metrics considered include accuracy (Acc), precision (Prec), recall (Rec), and F1-score, all derived from the confusion matrix. 

%

\vspace{-2mm}


\section{Evaluation Results}\label{evaluation_results}

Having provided the main foundations for our evaluation, we next show our detailed evaluation results to answer the aforementioned research questions.\vspace{-2mm}

\subsection{Overall Performance of Black-box AI Models} 

We begin by examining the overall performance of various black-box AI models in classifying anomalous AVs on both VeReMi and Sensor datasets. Tables~\ref{tbl:overall_performance_6_VeReMi} and \ref{tbl:overall_performance_10_sensor} summarize the key performance metrics—accuracy, precision, recall, and F1 score—collected for each model on both datasets. 

Table~\ref{tbl:overall_performance_6_VeReMi} demonstrates that, among the models tested on the VeReMi dataset, the Random Forest (RF) classifier achieves the best overall performance, with an accuracy of 0.80, precision of 0.83, recall of 0.88, and an F1 score of 0.86. The Decision Tree (DT) and K-Nearest Neighbors (KNN) models also perform relatively well, with comparable F1 scores of 0.84. The Deep Neural Network (DNN) and Support Vector Machine (SVM) models exhibit lower overall performance, although the DNN achieves the highest recall at 0.96. AdaBoost, while demonstrating solid recall at 0.91, shows lower precision and F1 scores compared to the random forest (RF) classifier. 

In contrast, Table~\ref{tbl:overall_performance_10_sensor} reveals that for the Sensor dataset, the AdaBoost model outperforms all other models with an impressive accuracy of 0.99, precision of 0.99, recall of 1.00, and an F1 score of 0.99. This is followed closely by the RF model, which also shows strong performance across all metrics, particularly in recall (0.98) and F1 score (0.94). The DNN model performs well with a balanced F1 score of 0.93, indicating reliable precision and recall. The KNN model excels in recall (0.97) but shows slightly lower precision and F1 scores compared to the top performers. The SVM model, while demonstrating robust performance, does not reach the same high levels as the other models, indicating a potential area for further optimization.

\begin{table}[h]
\centering
\caption{Overall performances for AI models with top 6 features for the VeReMi dataset.}
\vspace{-1mm}
\begin{tabular}{|c|c|c|c|c|}
\hline
\textbf{AI Model} & \textbf{Accuracy} & \textbf{Precision} & \textbf{Recall} & \textbf{F-1 score} \\ \hline
DT       & 0.79     & 0.82      & 0.87   & 0.84      \\ \hline
RF       & \textbf{0.80}     & \textbf{0.83}      & 0.88   & \textbf{0.86}      \\ \hline
DNN      & 0.66     & 0.67      & \textbf{0.96}   & 0.79      \\ \hline
KNN      & 0.78     & 0.82      & 0.85   & 0.84      \\ \hline
SVM      & 0.67     & 0.69      & 0.91   & 0.79      \\ \hline
AdaBoost & 0.73     & 0.74      &  0.91   & 0.82      \\ \hline
\end{tabular}
\label{tbl:overall_performance_6_VeReMi}
\vspace{-3mm}
\end{table}

\begin{table}[h]
\centering
\caption{Overall performances for AI models with top 10 features for the Sensor dataset.}
\vspace{-1mm}
\begin{tabular}{|c|c|c|c|c|}
\hline
\textbf{AI Model} & \textbf{Accuracy} & \textbf{Precision} & \textbf{Recall} & \textbf{F-1 score} \\ \hline
DT       & 0.85     & 0.89      & 0.92   & 0.90      \\ \hline
RF       & 0.90     & 0.90      & 0.98   & 0.94      \\ \hline
DNN      & 0.89     & 0.93      & 0.93   & 0.93      \\ \hline
KNN      & 0.84     & 0.85      & 0.97   & 0.91      \\ \hline
SVM      & 0.88     & 0.90      & 0.95   & 0.92      \\ \hline
AdaBoost & \textbf{0.99}     & \textbf{0.99}      & \textbf{1.00}   & \textbf{0.99}      \\ \hline
\end{tabular}
\label{tbl:overall_performance_10_sensor}
\vspace{-1mm}
\end{table}

\subsection{Ranking of Top Features for each XAI models}

Now, we provide the ranking of the top features for VeReMi and Sensor dataset for all of the three XAI methods considered in this work. These features are obtained from six AI models used in our framework both for binary and multiclass classification problems for anomaly detection.

\subsubsection{Binary Classification of VeReMi Dataset}

\textbf{Top Features for SHAP:}
We obtained six sets of top features for six of our AI models using SHAP for binary classification. Table~\ref{SHAP_AI_features} shows the list of top features for six different AI models elicited from SHAP. Among the features, "pos\_x" consistently ranks highest across most methods, while "pos\_z" consistently ranks lowest. Other features like "spd\_y" and "spd\_x" have varying ranks, indicating their differing importance across the methods.

\begin{table}
\centering
\vspace{2mm}
\caption{Ranking of VeReMi data features according to SHAP for anomaly detection binary classification.}

\vspace{-1mm}
\label{SHAP_AI_features}
\begin{tabular}{|c|c|c|c|c|c|c|c|} 
\hline
   { \textbf{Feature}}    & \textbf{DT} & \textbf{RF} & \textbf{DNN} & \textbf{KNN} & \textbf{SVM} & \textbf{AdaBoost}  \\ 
\hline
pos\_x & 1           & 1           & 1            & 3            & 2            & 3                 \\ 
\hline
pos\_y & 3           & 3           & 3            & 4            & 1            & 1                 \\ 
\hline
pos\_z & 6           & 6           & 6            & 6            & 6            & 6                 \\ 
\hline
spd\_x & 4           & 4           & 4            & 2            & 3            & 4                 \\ 
\hline
spd\_y & 2           & 2           & 2            & 1            & 4            & 2                 \\ 
\hline
spd\_z & 5           & 5           & 5            & 5            & 5            & 5                 \\
\hline
\end{tabular}
\vspace{-1mm}
\end{table}

\textbf{Top Features for LIME:} We now provide the six sets of top features for our six AI models extracted by LIME for binary classification. Table~\ref{LIME_AI_features} depicts that feature "spd\_y" is often considered the most significant across various methods, highlighting its importance. In contrast, "pos\_z" is persistently ranked the lowest, implying its lesser significance. The remaining features, including "pos\_x," "pos\_y," and "spd\_x," show fluctuating ranks, suggesting that their importance varies depending on the method used.

\begin{table}
\centering
\caption{Ranking of VeReMi data features according for LIME for anomaly detection binary classification.}

\vspace{-1mm}
\label{LIME_AI_features}
\begin{tabular}{|c|c|c|c|c|c|c|c|} 
\hline
   { \textbf{Feature}}    & \textbf{DT} & \textbf{RF} & \textbf{DNN} & \textbf{KNN} & \textbf{SVM} & \textbf{AdaBoost}  \\ 
\hline
pos\_x & 3           & 2           & 2            & 2            & 2            & 2                 \\ 
\hline
pos\_y & 4           & 4           & 1            & 4            & 3            & 3                 \\ 
\hline
pos\_z & 6           & 6           & 6            & 6            & 6            & 6                 \\ 
\hline
spd\_x & 2           & 3           & 3            & 3            & 1            & 4                 \\ 
\hline
spd\_y & 1           & 1           & 4            & 1            & 4            & 1                 \\ 
\hline
spd\_z & 5           & 5           & 5            & 5            & 5            & 5                 \\
\hline
\end{tabular}
\vspace{-3mm}
\end{table}

\textbf{Top Features for DALEX:} Similarly, following the above procedure, we present Table~\ref{DALEX_AI_features} shows the top ranked features for DALEX method for our six AI models for binary classification. The features "pos\_x" and "pos\_y" consistently have high rankings, showing their great value across the majority of models. The feature "spd\_x" also demonstrates its high ranking, indicating its significance in these procedures. On the other hand, the feature "pos\_z" regularly receives the lowest ranking, indicating that it is the least significant. The variables "spd\_y" and "spd\_z" are ranked as intermediate and low, respectively, showing their differing levels of significance across the models.

\begin{table}
\centering
\vspace{-2mm}
\caption{Ranking of VeReMi data features according to DALEX for anomaly detection binary classification.}
\vspace{-1mm}
\label{DALEX_AI_features}
\begin{tabular}{|c|c|c|c|c|c|c|c|} 
\hline
   { \textbf{Feature}}    & \textbf{DT} & \textbf{RF} & \textbf{DNN} & \textbf{KNN} & \textbf{SVM} & \textbf{AdaBoost}  \\ 
\hline
pos\_x & 1           & 1           & 2            & 2            & 3            & 1                 \\ 
\hline
pos\_y & 2           & 2           & 1            & 1            & 1            & 3                 \\ 
\hline
pos\_z & 6           & 6           & 6            & 6            & 6            & 6                 \\ 
\hline
spd\_x & 3           & 3           & 3            & 3            & 2            & 2                 \\ 
\hline
spd\_y & 4           & 4           & 4            & 4            & 4            & 4                 \\ 
\hline
spd\_z & 5           & 5           & 5            & 5            & 5            & 5                 \\
\hline
\end{tabular}
\vspace{-3mm}
\end{table}

\subsubsection{Binary Classification of Sensor Dataset}

\textbf{Top Features for SHAP:}
Through the same aforementioned process, we obtained that most AI algorithms consistently place features like "Lane Alignment" at the top position (Table~\ref{SHAP_sensor_features}), underscoring their high significance. The importance of other features, such as "Location" and "Frequency," fluctuates depending on the specific AI model used. Overall, the rankings provide insights into how each feature's relevance varies according to the different AI algorithms applied to the Sensor dataset.

\begin{table}[ht]
\centering
\vspace{2mm}
\caption{Ranking of main features in the Sensor dataset according to SHAP for the six different AI models.}
\vspace{-1mm}
\begin{tabular}{|c|c|c|c|c|c|c|} 
\hline
\textbf{Feature}   & \textbf{DT} & \textbf{RF} & \textbf{DNN} & \textbf{KNN} & \textbf{SVM} & \textbf{AdaBoost} \\ 
\hline
Formality          & 6           & 5           & 5            & 5            & 5            & 5                 \\ 
\hline
Location           & 10          & 10          & 10           & 10           & 8            & 10                \\ 
\hline
Frequency          & 8           & 6           & 8            & 9            & 9            & 8                 \\ 
\hline
Speed              & 7           & 8           & 6            & 4            & 6            & 6                 \\ 
\hline
Correlation        & 9           & 9           & 9            & 8            & 10           & 9                 \\ 
\hline
Lane Alignment     & 1           & 1           & 3            & 1            & 3            & 1                 \\ 
\hline
Headway Time       & 5           & 7           & 7            & 7            & 7            & 7                 \\ 
\hline
Protocol           & 4           & 4           & 4            & 6            & 4            & 2                 \\ 
\hline
Plausibility       & 2           & 3           & 2            & 3            & 2            & 3                 \\ 
\hline
Consistency        & 3           & 2           & 1            & 2            & 1            & 4                 \\
\hline
\end{tabular}
\label{SHAP_sensor_features}
\end{table}

\textbf{Top Features for LIME:}
Table~\ref{LIME_sensor_features} ranks Sensor dataset's features based on their importance according to LIME across the six AI methods. Key features like "Location" and "Consistency" show varying importance across different models, with "Location" ranking highly in Decision Tree, KNN, SVM, and AdaBoost, while "Consistency" ranks consistently high across all methods.

\begin{table}[t!]
\centering
\caption{Ranking of main features in the Sensor dataset according to LIME for the six different AI models.}
\vspace{-1mm}
\begin{tabular}{|c|c|c|c|c|c|c|} 
\hline
\textbf{Feature}   & \textbf{DT} & \textbf{RF} & \textbf{DNN} & \textbf{KNN} & \textbf{SVM} & \textbf{AdaBoost} \\ 
\hline
Formality          & 6           & 6           & 5            & 5            & 5            & 9                 \\ 
\hline
Location           & 1          & 2          & 10           & 1           & 2            & 1                \\ 
\hline
Frequency          & 10           & 8           & 8            & 8            & 10            & 10                 \\ 
\hline
Speed              & 7           & 7           & 6            & 9            & 7            & 6                 \\ 
\hline
Correlation        & 3           & 1           & 9            & 3            & 1           & 4                 \\ 
\hline
Lane Alignment     & 4           & 4           & 3            & 4            & 4            & 3                 \\ 
\hline
Headway Time       & 8           & 10           & 7            & 7            & 9            & 5                 \\ 
\hline
Protocol           & 5           & 5           & 4            & 6            & 6            & 8                 \\ 
\hline
Plausibility       & 9           & 9           & 2            & 10            & 8            & 7                 \\ 
\hline
Consistency        & 2           & 3           & 1            & 2            & 3            & 2                 \\
\hline
\end{tabular}
\label{LIME_sensor_features}
\end{table}

\textbf{Top Features for DALEX:}
We also utilized DALEX to rank various features according to their significance across several machine learning models, including DT, RF, DNN, KNN, SVM, AdaBoost. Table~\ref{DALEX_sensor_features} suggests notable findings indicate that "Protocol" holds the highest importance for DNN, while "Location" is crucial for KNN and SVM. On the other hand, "Frequency" is found to have minimal influence across all the models.

\begin{table}[t]
\centering
\vspace{2mm}
\caption{Ranking of the main features in the Sensor dataset according to DALEX for the six different AI models.}
\vspace{-1mm}
\begin{tabular}{|c|c|c|c|c|c|c|} 
\hline
\textbf{Feature}   & \textbf{DT} & \textbf{RF} & \textbf{DNN} & \textbf{KNN} & \textbf{SVM} & \textbf{AdaBoost} \\ 
\hline
Formality          & 5           & 6           & 6            & 6            & 6            & 5                 \\ 
\hline
Location           & 3          & 3          & 1           & 1           & 2            & 6                \\ 
\hline
Frequency          & 8           & 8           & 10            & 10            & 10            & 10                 \\ 
\hline
Speed              & 7           & 7           & 8            & 8            & 7            & 8                 \\ 
\hline
Correlation        & 2           & 1           & 2            & 2            & 1           & 3                 \\ 
\hline
Lane Alignment     & 4           & 2           & 4            & 4            & 3            & 1                 \\ 
\hline
Headway Time       & 9           & 9           & 7            & 7            & 9            & 7                 \\ 
\hline
Protocol           & 1           & 5           & 5            & 5            & 5            & 2                 \\ 
\hline
Plausibility       & 10           & 10           & 9            & 9            & 8            & 4                 \\ 
\hline
Consistency        & 6           & 4           & 3            & 3            & 4            & 9                 \\
\hline
\end{tabular}
\label{DALEX_sensor_features}
\end{table}

Having shown individual feature importance for each XAI method and each AI model for binary classification problem, we next show such individual feature importance for the multiclass classification problem for VeReMi dataset.

\subsubsection{Multiclass Classification of VeReMi Dataset}
\textbf{Top Features for SHAP:} Using SHAP values, Table~\ref{SHAP_AI_features_multiclass} ranks the features from VeReMi data based on their influence in multiclass classification setup for anomaly detection across various models. Features such as "pos\_x" and "pos\_y" consistently demonstrate high significance across these models, whereas "spd\_z" generally exhibits lower impact.

\begin{table}
\centering
\caption{Ranking of main features for VeReMi dataset according to SHAP for multiclass classification setup.}
\vspace{-1mm}
\label{SHAP_AI_features_multiclass}
\begin{tabular}{|c|c|c|c|c|c|c|c|} 
\hline
   { \textbf{Feature}}    & \textbf{DT} & \textbf{RF} & \textbf{DNN} & \textbf{KNN} & \textbf{SVM} & \textbf{AdaBoost}  \\ 
\hline
pos\_x & 1           & 1           & 2            & 2            & 3            & 1                 \\ 
\hline
pos\_y & 2           & 2           & 1            & 1            & 4            & 2                 \\ 
\hline
pos\_z & 6           & 6           & 6            & 3            & 2            & 3                 \\ 
\hline
spd\_x & 4           & 4           & 3            & 4            & 1            & 4                 \\ 
\hline
spd\_y & 3           & 3           & 4            & 5            & 6            & 6                 \\ 
\hline
spd\_z & 5           & 5           & 5            & 6            & 5            & 5                 \\
\hline
\end{tabular}
\vspace{-3mm}
\end{table}

\textbf{Top Features for LIME:} Table~\ref{LIME_AI_features_multiclass} presents a hierarchical arrangement of features from the VeReMi dataset, based on their significance in multiclass classification tasks. The ranking is derived from LIME analysis applied to a diverse set of machine learning models. Notably, "pos\_x" consistently ranks as the most important feature across all models. "pos\_y" also shows significant importance, especially for SVM and AdaBoost models. "spd\_z" consistently ranks lowest in importance across all models evaluated by LIME.

\begin{table}[t!]
\centering
\vspace{1mm}
\caption{Ranking of main features for VeReMi dataset  according to LIME for multiclass classification setup.}
\vspace{-1mm}
\label{LIME_AI_features_multiclass}
\begin{tabular}{|c|c|c|c|c|c|c|c|} 
\hline
   { \textbf{Feature}}    & \textbf{DT} & \textbf{RF} & \textbf{DNN} & \textbf{KNN} & \textbf{SVM} & \textbf{AdaBoost}  \\ 
\hline
pos\_x & 1           & 2           & 2            & 2            & 3            & 1                 \\ 
\hline
pos\_y & 4           & 3           & 1            & 3            & 4            & 3                 \\ 
\hline
pos\_z & 6           & 6           & 6            & 1            & 2            & 4                 \\ 
\hline
spd\_x & 3           & 4           & 3            & 4            & 1            & 2                 \\ 
\hline
spd\_y & 2           & 1           & 4            & 5            & 5            & 5                 \\ 
\hline
spd\_z & 5           & 5           & 5            & 6            & 6            & 6                 \\
\hline
\end{tabular}
\vspace{-1mm}
\end{table}

\textbf{Top Features DALEX:} Table~\ref{DALEX_AI_features_multiclass} ranks features from VeReMi data according to their significance for multiclass classification using DALEX across the six different AI models, "spd\_y" is consistently the most critical attribute in all models. "pos\_x" and "pos\_y" are also of considerable significance in the majority of models, while "pos\_z" is generally of lower importance. The importance of "spd\_z" is consistently the lowest among all models using DALEX.

In the above segment of our work, we have identified the top features using the aforementioned XAI methods (SHAP, LIME, DALEX) and AI models. We next perform feature ensemble (Section~\ref{sec:framwork}) to derive a unified set of features for each of our three cases: binary classification for the VeReMi dataset, binary classification for the Sensor dataset, and multiclass classification for the VeReMi dataset.

\begin{table}[t!]
\centering
\caption{Ranking of main features of VeReMi dataset according to DALEX for multiclass classification setup.}
\vspace{-1mm}
\label{DALEX_AI_features_multiclass}
\begin{tabular}{|c|c|c|c|c|c|c|c|} 
\hline
   { \textbf{Feature}}    & \textbf{DT} & \textbf{RF} & \textbf{DNN} & \textbf{KNN} & \textbf{SVM} & \textbf{AdaBoost}  \\ 
\hline
pos\_x & 4           & 3           & 2            & 3            & 3            & 4                 \\ 
\hline
pos\_y & 3           & 4           & 1            & 4            & 2            & 3                 \\ 
\hline
pos\_z & 6           & 6           & 6            & 1            & 2            & 4                 \\ 
\hline
spd\_x & 2           & 2           & 3            & 1            & 4            & 2                 \\ 
\hline
spd\_y & 1           & 1           & 4            & 2            & 1            & 1                 \\ 
\hline
spd\_z & 5           & 5           & 5            & 6            & 6            & 6                 \\
\hline
\end{tabular}
\vspace{-3mm}
\end{table}

\begin{table*}[t!]
\centering
\caption{Top features for three XAI methods (SHAP, LIME, and DALEX) and the feature ensemble method (Leveled) for different datasets and anomaly detection setups (VeReMi Binary, Sensor, and VeReMi Multiclass classification).}
\label{combined_features}
\resizebox{\linewidth}{!}{
\begin{tabular}{|c|c|c|c|c|}
\hline
\textbf{Dataset}                    & \textbf{SHAP}        & \textbf{LIME}          & \textbf{DALEX}         & \textbf{Leveled}       \\ \hline

\textbf{VeReMi Binary}              & pos\_x               & spd\_y                 & pos\_x                 & pos\_x                 \\ 
                                    & pos\_y               & pos\_x                 & pos\_y                 & pos\_y                 \\ 
                                    & spd\_x               & spd\_x                 & spd\_x                 & spd\_x                 \\ 
                                    & spd\_y               & pos\_y                 & spd\_y                 & spd\_y                 \\ \hline

\textbf{Sensor}                     & Location            & Location             & Location             & Location             \\ 
                                    & Frequency           & Correlation          & Lane Alignment       & Lane Alignment       \\ 
                                    & Lane Alignment      & Consistency          & Correlation          & Consistency          \\ 
                                    & Protocol            & Lane Alignment       & Consistency          & Correlation          \\ 
                                    & Consistency         & Protocol             & Formality            & Protocol             \\ \hline

\textbf{VeReMi Multiclass}          & pos\_x               & pos\_x                 & spd\_y                 & pos\_x                 \\ 
                                    & pos\_y               & pos\_y                 & spd\_x                 & pos\_y                 \\ 
                                    & spd\_x               & spd\_x                 & pos\_x                 & spd\_x                 \\ 
                                    & pos\_z               & pos\_z                 & pos\_y                 & spd\_y                 \\ \hline
\end{tabular}
}
\end{table*}

\subsection{Fusion of Features and their performance on Independent Classifiers}

We will first evaluate the aforementioned three setups using feature sets generate by our feature ensemble approach. The consolidated feature sets are subsequently used to train three independent classifiers: CatBoost, LGBM, and LR. We then compare the performance of these classifiers using the fused feature sets against their performance when trained on the raw top features identified by each XAI method individually.

\subsubsection{VeReMi Binary Class Classification}
Recall, we do the leveled-feature fusion in three phases. First, we get three sets of features for each of our XAI methods (SHAP, LIME, and DALEX) for each of our six AI models. Secondly, for each AI model we fuse the three sets features (obtained from the first phase) from the three XAI methods and get one set of features for each of the six AI models. In the end, we combine the six features from six AI models into a single set of fused features depending on their frequency.

We now feed the set top features from fusion of features for just SHAP, LIME, and DALEX from all the six models to three of our independent classifiers and observe the comparison of performance against the independent models fed with top features from leveled-feature fusion technique. We emphasize that for VeReMi dataset we focus on top-$4$ features. The top-$4$ features we used for the experiment are shown in VeReMi binary classification in Table~\ref{combined_features}. ``pos\_x,'' ``pos\_y,'' and ``spd\_x'' are the overall top features for VeReMi dataset (in both binary and multiclass setups) while ``Location'' and ``Lane Alignment,' and ``Consistency'' are the overall top features for Sensor dataset.

VeReMi binary classification in Table~\ref{comparison_combined} shows that the CatBoost classifier maintains consistent performance across all methods (SHAP, LIME, DALEX, and Leveled) with an accuracy of 0.82, precision of 0.86, recall around 0.91-0.92, and F1-score of 0.89. LGBM and Logistic Regression also show minor variations across methods, but generally, the novel feature fusion method (Leveled) performs comparably to individual XAI methods, indicating that Leveled features are effective in maintaining classifier performance.

\subsubsection{Binary Classification of Sensor Dataset}

In the next phase of our analysis, we utilize the top features derived from the fusion of SHAP, LIME, and DALEX outputs across all six models. These consolidated feature sets are then fed into three independent classifiers. We compare the performance of these classifiers against the same models when trained on features selected through our leveled-feature fusion technique. It is important to note that for the Sensor dataset, our focus is specifically on the top-$5$ features. The specific set of top-$5$ features employed in this experimental setup is detailed in Sensor classification in Table~\ref{combined_features}. This approach allows us to evaluate the efficacy of our feature fusion methodology in the context of the Sensor dataset.

Sensor classification in Table~\ref{comparison_combined} compares the performance of CatBoost, LGBM, and Logistic Regression classifiers on the Sensor dataset employing SHAP, LIME, DALEX, and a novel feature ensemble method (Leveled) for binary classification. Across all classifiers, the Leveled technique consistently produces competitive results, frequently matching or slightly beating specific XAI methods. Notably, the CatBoost classifier performs well with Leveled features, with an accuracy of 0.82, precision of 0.86, recall of 0.92, and F1-score of 0.89, demonstrating the Leveled method's efficiency in improving classification performance.

\begin{table*}[t!]
\centering
\caption{Comparison of results of three independent classifiers (CatBoost, LGBM, and LR) for our XAI methods (SHAP, LIME, DALEX) and novel feature fusion method (Leveled) for VeReMi binary classification, Sensor binary classification, and VeReMi multiclass classification. The best result is highlighted in \textbf{bold} for each metric (Acc, Prec, Rec, and F-1) for each AI model (CatBoost, LGBM, LR) within each dataset (VeReMi Binary, Sensor Binary, and VeReMi Multiclass).}
\label{comparison_combined}
\resizebox{\columnwidth}{!}{
\begin{tabular}{|c|cccc|cccc|cccc|}
\hline
\textbf{Metrics} & \multicolumn{4}{c|}{\textbf{CatBoost}}                                                                                           & \multicolumn{4}{c|}{\textbf{LGBM}}                                                                                               & \multicolumn{4}{c|}{\textbf{LR}}                                                                                                 \\ \hline
        & \multicolumn{1}{c|}{\textbf{SHAP}} & \multicolumn{1}{c|}{\textbf{LIME}} & \multicolumn{1}{c|}{\textbf{DALEX}} & \textbf{Leveled} & \multicolumn{1}{c|}{\textbf{SHAP}} & \multicolumn{1}{c|}{\textbf{LIME}} & \multicolumn{1}{c|}{\textbf{DALEX}} & \textbf{Leveled} & \multicolumn{1}{c|}{\textbf{SHAP}} & \multicolumn{1}{c|}{\textbf{LIME}} & \multicolumn{1}{c|}{\textbf{DALEX}} & \textbf{Leveled} \\ \hline

\multicolumn{13}{|c|}{\textbf{VeReMi Binary Classification}} \\ \hline
Acc              & \multicolumn{1}{c|}{\textbf{0.82}} & \multicolumn{1}{c|}{\textbf{0.82}} & \multicolumn{1}{c|}{\textbf{0.82}} & \textbf{0.82}    & \multicolumn{1}{c|}{0.79}          & \multicolumn{1}{c|}{0.80}          & \multicolumn{1}{c|}{0.80}          & \textbf{0.80}    & \multicolumn{1}{c|}{\textbf{0.80}}          & \multicolumn{1}{c|}{\textbf{0.80}}          & \multicolumn{1}{c|}{\textbf{0.80}}          & \textbf{0.80}    \\ \hline
Prec             & \multicolumn{1}{c|}{\textbf{0.86}} & \multicolumn{1}{c|}{\textbf{0.86}} & \multicolumn{1}{c|}{\textbf{0.86}} & \textbf{0.86}    & \multicolumn{1}{c|}{\textbf{0.85}} & \multicolumn{1}{c|}{0.84}          & \multicolumn{1}{c|}{0.84}          & \textbf{0.84}    & \multicolumn{1}{c|}{0.82}          & \multicolumn{1}{c|}{0.82}          & \multicolumn{1}{c|}{0.82}          & \textbf{0.82}    \\ \hline
Rec              & \multicolumn{1}{c|}{\textbf{0.92}} & \multicolumn{1}{c|}{0.91}          & \multicolumn{1}{c|}{0.91}          & \textbf{0.92}    & \multicolumn{1}{c|}{0.88}          & \multicolumn{1}{c|}{\textbf{0.90}} & \multicolumn{1}{c|}{\textbf{0.90}} & \textbf{0.90}    & \multicolumn{1}{c|}{\textbf{0.94}}          & \multicolumn{1}{c|}{\textbf{0.95}}          & \multicolumn{1}{c|}{\textbf{0.95}} & \textbf{0.95}    \\ \hline
F-1              & \multicolumn{1}{c|}{\textbf{0.89}} & \multicolumn{1}{c|}{\textbf{0.89}} & \multicolumn{1}{c|}{\textbf{0.89}} & \textbf{0.89}    & \multicolumn{1}{c|}{0.86}          & \multicolumn{1}{c|}{\textbf{0.87}} & \multicolumn{1}{c|}{\textbf{0.87}} & \textbf{0.87}    & \multicolumn{1}{c|}{0.88}          & \multicolumn{1}{c|}{0.88}          & \multicolumn{1}{c|}{0.88}          & \textbf{0.88}    \\ \hline

\multicolumn{13}{|c|}{\textbf{Sensor Classification}} \\ \hline
Acc              & \multicolumn{1}{c|}{\textbf{0.82}} & \multicolumn{1}{c|}{0.80}          & \multicolumn{1}{c|}{0.81}          & \textbf{0.82}    & \multicolumn{1}{c|}{\textbf{0.79}} & \multicolumn{1}{c|}{0.76}          & \multicolumn{1}{c|}{0.78}          & 0.79             & \multicolumn{1}{c|}{\textbf{0.80}} & \multicolumn{1}{c|}{0.79}          & \multicolumn{1}{c|}{0.79}          & \textbf{0.80}             \\ \hline
Prec             & \multicolumn{1}{c|}{\textbf{0.86}} & \multicolumn{1}{c|}{0.84}          & \multicolumn{1}{c|}{0.81}          & \textbf{0.86}    & \multicolumn{1}{c|}{\textbf{0.84}} & \multicolumn{1}{c|}{0.82}          & \multicolumn{1}{c|}{\textbf{0.84}} & 0.84             & \multicolumn{1}{c|}{\textbf{0.82}} & \multicolumn{1}{c|}{0.81}          & \multicolumn{1}{c|}{0.81}          & \textbf{0.82}             \\ \hline
Rec              & \multicolumn{1}{c|}{\textbf{0.92}} & \multicolumn{1}{c|}{0.90}          & \multicolumn{1}{c|}{0.86}          & \textbf{0.92}    & \multicolumn{1}{c|}{0.89}          & \multicolumn{1}{c|}{0.86}          & \multicolumn{1}{c|}{\textbf{0.90}} & \textbf{0.90}    & \multicolumn{1}{c|}{\textbf{0.95}} & \multicolumn{1}{c|}{0.92}          & \multicolumn{1}{c|}{0.91}          & \textbf{0.95}             \\ \hline
F-1              & \multicolumn{1}{c|}{\textbf{0.89}} & \multicolumn{1}{c|}{0.87}          & \multicolumn{1}{c|}{0.84}          & \textbf{0.89}    & \multicolumn{1}{c|}{0.84}          & \multicolumn{1}{c|}{0.84}          & \multicolumn{1}{c|}{\textbf{0.87}} & 0.87             & \multicolumn{1}{c|}{\textbf{0.88}} & \multicolumn{1}{c|}{0.85}          & \multicolumn{1}{c|}{0.86}          & \textbf{0.88}             \\ \hline

\multicolumn{13}{|c|}{\textbf{VeReMi Multiclass Classification}} \\ \hline
Acc              & \multicolumn{1}{c|}{\textbf{0.67}} & \multicolumn{1}{c|}{\textbf{0.67}} & \multicolumn{1}{c|}{\textbf{0.67}} & \textbf{0.67}    & \multicolumn{1}{c|}{\textbf{0.67}} & \multicolumn{1}{c|}{\textbf{0.67}} & \multicolumn{1}{c|}{\textbf{0.67}} & 0.64             & \multicolumn{1}{c|}{\textbf{0.67}} & \multicolumn{1}{c|}{\textbf{0.67}} & \multicolumn{1}{c|}{\textbf{0.67}} & \textbf{0.67}             \\ \hline
Prec             & \multicolumn{1}{c|}{\textbf{0.67}} & \multicolumn{1}{c|}{\textbf{0.67}} & \multicolumn{1}{c|}{\textbf{0.67}} & \textbf{0.67}    & \multicolumn{1}{c|}{0.67} & \multicolumn{1}{c|}{0.67} & \multicolumn{1}{c|}{0.67} & \textbf{0.73}             & \multicolumn{1}{c|}{\textbf{0.67}} & \multicolumn{1}{c|}{\textbf{0.67}} & \multicolumn{1}{c|}{\textbf{0.67}} & \textbf{0.67}             \\ \hline
Rec              & \multicolumn{1}{c|}{\textbf{1.00}} & \multicolumn{1}{c|}{\textbf{1.00}} & \multicolumn{1}{c|}{\textbf{1.00}} & \textbf{1.00}    & \multicolumn{1}{c|}{\textbf{1.00}} & \multicolumn{1}{c|}{\textbf{1.00}} & \multicolumn{1}{c|}{\textbf{1.00}} & 0.93             & \multicolumn{1}{c|}{\textbf{1.00}} & \multicolumn{1}{c|}{\textbf{1.00}} & \multicolumn{1}{c|}{\textbf{1.00}} & \textbf{1.00}             \\ \hline
F-1              & \multicolumn{1}{c|}{\textbf{0.80}} & \multicolumn{1}{c|}{\textbf{0.80}} & \multicolumn{1}{c|}{\textbf{0.80}} & \textbf{0.80}    & \multicolumn{1}{c|}{0.80} & \multicolumn{1}{c|}{0.80} & \multicolumn{1}{c|}{0.80} & \textbf{0.82}             & \multicolumn{1}{c|}{\textbf{0.80}} & \multicolumn{1}{c|}{\textbf{0.80}} & \multicolumn{1}{c|}{\textbf{0.80}} & \textbf{0.80}             \\ \hline

\end{tabular}
}
\vspace{-2mm}
\end{table*}

\subsubsection{Multiclass Classification of VeReMi Dataset:}

Next, we take into account the multiclass classification for VeReMi dataset and we compare the results of our independent classifiers on different sets of top-$4$ features following the aforementioned procedures. The top-$4$ features for VeReMi dataset for multiclass which are fed to the independent classifiers are shown in VeReMi multiclass classification in Table~\ref{combined_features}.

VeReMi multiclass classification in Table~\ref{comparison_combined} presents a comparison of the performance of CatBoost, LGBM, and Logistic Regression classifiers for VeReMi multiclass classification. The evaluation is done using SHAP, LIME, DALEX, and a novel feature ensemble method called Leveled. All classifiers consistently attain similar levels of accuracy and precision, with values around 0.67 for accuracy and precision, and 0.80 for F1-score. The recall rate is continuously high at 1.00 for most approaches, except for LGBM with Leveled features, which has a slightly lower recall rate of 0.93. The Leveled technique often achieves comparable performance to the various XAI methods, with a modest enhancement in F1-score observed for LGBM.

%% file: Discussion.tex

\section{Limitations and Discussion}\label{sec: Discussion}

\subsection{Limitations}
%
%


While our proposed XAI-based feature ensemble framework demonstrates promising results for enhancing anomaly detection in autonomous driving systems, it is important to acknowledge several limitations in our current study and discuss their implications along with future enhancements.

\textbf{Dataset Limitations:} Our study primarily relied on two datasets - VeReMi and Sensor. While these datasets are widely used in the field, they may not fully represent the complexity and diversity of different real-world autonomous driving scenarios. The VeReMi dataset, being simulation-based, may not capture all the nuances of real-world vehicular networks. Future work should validate our approach on more diverse and extensive real-world datasets to ensure generalizability. 
For instance, there are other autonomous driving
datasets (e.g., A2D2~\cite{geyer2020a2d2}, and Pass~\cite{hu2022pass}) with other features
that are not considered in our studied datasets.

\textbf{Feature Selection Constraints:} We focused on a limited number of top features (four for VeReMi and five for Sensor datasets) to evaluate our feature ensemble approach. While this allowed for computational efficiency, it may have excluded potentially valuable features that could contribute to anomaly detection. A more comprehensive analysis of feature interactions and their collective impact on model performance could provide deeper insights.

\textbf{Temporal Aspects and Model Diversity:} Our current approach does not explicitly account for the temporal nature of autonomous driving data. Incorporating time-series analysis techniques or recurrent neural network architectures could potentially enhance the detection of time-dependent anomalies. Furthermore, although we employed a variety of AI models, our selection was not very exhaustive. The performance of our framework on other advanced models, such as Graph Neural Networks or Transformer-based models, which have shown promise in sequential (temporal) data analysis, remains unexplored. Expanding the range of models could potentially uncover additional insights.

%



\subsection{Discussion}

Despite the identified limitations, our XAI-based feature ensemble framework offers notable advantages and paves the way for new research avenues. The consistent performance improvements across various classifiers indicate that our approach effectively captures essential features for anomaly detection for autonomous driving systems. 

By integrating insights from multiple XAI methods, we achieve a better understanding of feature importance (via the features identified by frequency analysis of the features identified from the different XAI methods), potentially reducing biases inherent in individual techniques\cite{das2020opportunities}. The framework's ability to maintain or slightly enhance performance while improving interpretability is particularly significant in the context of autonomous driving, where both accuracy and explainability are crucial. 

Our results emphasize the value of combining multiple AI models and XAI methods, reflecting the complexity of anomaly detection and the need for multi-faceted approaches. Future research could explore unsupervised anomaly detection methods, incorporate domain knowledge into the feature ensemble (or fusion) process, and develop advanced feature ensemble techniques to capture non-linear feature interactions. Additionally, assessing the framework's performance on multi-modal data (e.g., combining sensor data with visual inputs) could enhance the evaluation of anomaly detection in autonomous driving systems.

%% file: Conclusion.tex

\section{Conclusion}\label{sec: conclusion}



This paper introduced a novel XAI-based feature ensemble framework that enhances anomaly detection in autonomous driving systems by integrating features from multiple XAI methods (SHAP, LIME, and DALEX) with various AI models to develop a feature fusion technique. Evaluated on the VeReMi and Sensor datasets, our approach consistently matches or outperforms individual XAI methods across different independent classifiers (CatBoost, LGBM, and Logistic Regression). Key findings include robust anomaly indicators from fused features, high performance in both binary and multiclass classification, and improved interpretability essential for stakeholder trust and regulatory compliance. While limitations such as dataset constraints and computational overhead exist, our work lays the foundation for future research in using XAI-based feature ensemble for enhancing understanding of anomaly detection process for autonomous driving. The main related future avenues for research include real-world validation, temporal analysis, and adversarial robustness. This research advances the development of more reliable, and secure autonomous driving systems, bridging the gap between high-performance anomaly detection and feature analysis using explainable AI.

\section*{Acknowledgment}
This work was supported by Lilly Endowment (AnalytixIN); and Enhanced Mentoring Program with Opportunities for
Ways to Excel in Research (EMPOWER) Grant from the Office of the Vice Chancellor for Research at IUPUI.